\documentclass[runningheads]{llncs}
\usepackage{graphicx}
\usepackage[nolist]{acronym}
\usepackage[perpage]{footmisc}

\begin{document}
\title{A Study on Dialog Act Recognition using Character-Level Tokenization\thanks{This work was supported by national funds through \ac{FCT} with reference UID/CEC/50021/2013 and by Universidade de Lisboa.}}
\titlerunning{Character-Level Dialog Act Recognition}  
%
\author{Eug\'{e}nio Ribeiro\inst{1,2} \and Ricardo Ribeiro\inst{1,3} \and David Martins de Matos\inst{1,2}}
\authorrunning{E. Ribeiro et al.}
%
\institute{L$^2$F {--} Spoken Language Systems Laboratory {--} INESC-ID Lisboa
\and
Instituto Superior T\'{e}cnico, Universidade de Lisboa, Portugal
\and
Instituto Universit\'{a}rio de Lisboa (ISCTE-IUL) \\
\email{eugenio.ribeiro@l2f.inesc-id.pt}}
\maketitle              
%
%
%

\begin{acronym}
	\acro{CNN}{Convolutional Neural Network}
    \acro{DNN}{Deep Neural Network}
    \acro{DRLM}{Discourse Relation Language Model}
    \acro{FCT}{Funda\c{c}\~{a}o para a Ci\^{e}ncia e a Tecnologia}
    \acro{GloVe}{Global Vectors for Word Representation}
    \acro{HMM}{Hidden Markov Model}
    \acro{LSTM}{Long Short-Term Memory}
    \acro{NLP}{Natural Language Processing}
    \acro{POS}{Part-of-Speech}
    \acro{RNN}{Recurrent Neural Network}
    \acro{RNNLM}{Recurrent Neural Network Language Model}
    \acro{SVM}{Support Vector Machine}
    \acro{WoZ}{Wizard of Oz}
\end{acronym}

%
%
%
%

\begin{abstract}
Dialog act recognition is an important step for dialog systems since it reveals the intention behind the uttered words. Most approaches on the task use word-level tokenization. In contrast, this paper explores the use of character-level tokenization. This is relevant since there is information at the sub-word level that is related to the function of the words and, thus, their intention. We also explore the use of different context windows around each token, which are able to capture important elements, such as affixes. Furthermore, we assess the importance of punctuation and capitalization. We performed experiments on both the Switchboard Dialog Act Corpus and the DIHANA Corpus. In both cases, the experiments not only show that character-level tokenization leads to better performance than the typical word-level approaches, but also that both approaches are able to capture complementary information. Thus, the best results are achieved by combining tokenization at both levels.
\end{abstract}

%
%
%
%
%

\section{Introduction}
\label{sec:introduction}

Dialog act recognition is important in the context of a dialog system, since it reveals the intention behind the words uttered by its conversational partners~\cite{Searle1969}. Knowing that intention allows the system to apply specialized interpretation strategies, accordingly. Recently, most approaches on dialog act recognition focus on applying different \ac{DNN} architectures to generate segment representations from word embeddings and combine them with context information from the surrounding segments~\cite{Kalchbrenner2013,Lee2016,Ji2016,Liu2017}. However, all of these approaches look at the segment at the word level. That is, they consider that a segment is a sequence of words and that its intention is revealed by the combination of those words. However, there are also cues for intention at the sub-word level. These cues are mostly related to the morphology of words. For instance, there are cases, such as adverbs of manner and negatives, in which the function, and hence the intention, of a word is related to its affixes. On the other hand, there are cases in which considering multiple forms of the same lexeme independently does not provide additional information concerning intention and the lemma suffices. Thus, it is interesting to explore dialog act recognition approaches that are able to capture this kind of information. In this paper, we explore the use of character-level tokenization with different context windows surrounding each token. Although character-level approaches are typically used for word-level classification tasks, such as \ac{POS} tagging~\cite{Santos2014}, they have also achieved interesting results on short-text classification tasks, such as language identification~\cite{Jaech2016} and review rating~\cite{Zhang2015}. In addition to the aspects concerning morphological information, using character-level tokenization allows us to assess the importance of aspects such as capitalization and punctuation. Additionally, we assess whether the obtained information can be combined with that obtained using word-level tokenization to improve the performance on the task. In this sense, in order to widen the scope of our conclusions, we performed experiments on two corpora, the Switchboard Dialog Act Corpus~\cite{Jurafsky1997} and DIHANA~\cite{Benedi2006}, which have different characteristics, including domain, the nature of the participants, and language {--} English and Spanish, respectively.

In the remainder of this paper we start by providing an overview of previous approaches on dialog act recognition, in Section~\ref{sec:related}. Then, in Section~\ref{sec:character}, we discuss why using character-level tokenization is relevant for the task. Section~\ref{sec:setup} describes our experimental setup, including the used datasets, classification approach, and word-level baselines. The results of our experiments are presented and discussed in Section~\ref{sec:results}. Finally, Section~\ref{sec:conclusions} states the most important conclusions of this study and provides pointers for future work.

%
%
%
%

\section{Related Work}
\label{sec:related}

Automatic dialog act recognition is a task that has been widely explored over the years, using multiple machine learning approaches, from \acp{HMM}~\cite{Stolcke2000} to \acp{SVM}~\cite{Gamback2011}. The article by Kr\'{a}l and Cerisara~\cite{Kral2010} provides an interesting overview of most of those approaches on the task. However, recently, similarly to many other \ac{NLP} tasks~\cite{Manning2015,Goldberg2016}, most approaches on dialog act recognition take advantage of different \ac{DNN} architectures.

To our knowledge, the first of those approaches was that by Kalchbrenner and Blunsom~\cite{Kalchbrenner2013}. They used a \ac{CNN}-based approach to generate segment representations from randomly initialized 25-dimensional word embeddings and a \ac{RNN}-based discourse model to combine the sequence of segment representations with speaker information and output the corresponding sequence of dialog acts.

Lee and Dernoncourt~\cite{Lee2016} compared the performance of a \ac{LSTM} unit against that of a \ac{CNN} to generate segment representations from 200-dimensional \ac{GloVe} embeddings~\cite{Pennington2014} pre-trained on Twitter data. Those segment representations were then fed to a 2-layer feed-forward network that combined them with context information from the preceding segments. The best results were obtained using the \ac{CNN}-based approach combined with information from two preceding segments in the form of their representation.

Ji et al.~\cite{Ji2016} used a \ac{DRLM} with a hybrid architecture that combined a \ac{RNNLM}~\cite{Mikolov2010} with a latent variable model over shallow discourse structure. This way, the model can learn vector representations trained discriminatively, while maintaining a probabilistic representation of the targeted linguistic element which, in this context, is the dialog act. In order to function as a classifier, the model was trained to maximize the conditional probability of a sequence of dialog acts given a sequence of segments.

The previous studies explored the use of a single recurrent or convolutional layer. However, the top performing approaches use multiple of those layers. On the one hand, Khanpour et al.~\cite{Khanpour2016} achieved their best results by combining the outputs of a stack of 10 \ac{LSTM} units, in order to capture long distance relations between tokens. On the other hand, Liu et al.~\cite{Liu2017} combined the outputs of three parallel \acp{CNN} with different context window sizes, in order to capture different functional patterns. Both studies used Word2Vec~\cite{Mikolov2013} embeddings as input to the network. However, their dimensionality and training data varied.

Additionally, Liu et al.~\cite{Liu2017} explored the use of context information concerning speaker changes and from the surrounding segments. Concerning the latter, they used approaches that relied on discourse models, as well as others that combined the context information directly with the segment representation. Similarly to our previous study using \acp{SVM}~\cite{Ribeiro2015}, they concluded that providing that information in the form of the classification of the surrounding segments leads to better results than using their words. Furthermore, both studies have shown that the first preceding segment is the most important and that the influence decays with the distance.

%
%
%
%

\section{Character-Level Tokenization}
\label{sec:character}

It is interesting to explore character-level tokenization because it allows us to capture morphological information that is at the sub-word level and, thus, cannot be directly captured using word-level tokenization. Considering the task at hand, that information is relevant since it may provide cues for identifying the intention behind the words. When someone selects a set of words to form a segment that transmits a certain intention, each of those words is typically selected because it has a function that contributes to that transmission. In this sense, affixes are tightly related to word function, especially in fusional languages. Thus, the presence of certain affixes is a cue for intention, independently of the lemma. However, there are also cases, such as when affixes are used for subject-verb agreement, in which the cue for intention is in the lemmas and, thus, considering multiple forms of the same lexeme does not provide additional information.

Information concerning lemmas and affixes cannot be captured from single independent characters. Thus, it is necessary to consider the context surrounding each token and look at groups of characters. The size of the context window plays an important part in what information can be captured. For instance, English affixes are typically short, but in other languages, such as Spanish, there are longer commonly used affixes. Furthermore, to capture the lemmas of long words, and even inter-word relations, wider context window sizes must be considered. However, using wide context windows impairs the ability to capture information from short groups of characters, as additional irrelevant characters are considered. This suggests that, in order to capture all the relevant information, multiple context windows should be used.

Using character-level tokenization also allows us to consider punctuation, which is able to provide both direct and indirect cues for dialog act recognition. For instance, an interrogation mark provides a direct cue that the intention is related to knowledge seeking. On the other hand, commas structure the segment, indirectly contributing to the transmission of an intention.

Additionally, character-level tokenization allows us to consider capitalization information. However, in the beginning of a segment, capitalization only signals that beginning and, thus, considering it only introduces entropy. In the middle of a segment, capitalization is typically only used to distinguish proper nouns, which are not related to intention. Thus, capitalization information is not expected to contribute to the task.

Finally, note that previous studies have shown that word-level information is relevant for the task. In this sense, it is interesting to assess whether that information can be captured using character-level tokenization or if there are specific aspects that require specialized approaches.

%
%
%
%

\section{Experimental Setup}
\label{sec:setup}

In order to assess the validity of the hypotheses proposed in the previous section, we performed experiments on different corpora and compared the performance of word- and character-level tokenization. The used datasets, classification approach, and word-level baselines are described below. 

\subsection{Datasets}

In order to widen the scope of the conclusions drawn in the study, we selected two corpora with different characteristics to perform our experiments on. On the one hand, the Switchboard Dialog Act Corpus~\cite{Jurafsky1997}, henceforth referred to as Switchboard, is the most explored corpus for dialog act recognition. It features 1,155 manually transcribed human-human dialogs in English, with variable domain, containing 223,606 segments. The set is partitioned into a training set of 1,115 conversations, a test set of 19 conversations, and a future use set of 21 conversations~\cite{Stolcke2000}. In our experiments, we used the latter as a validation set. In terms of dialog act annotations, we used the most used version of its tag set, which features 42 domain-independent labels.

On the other hand, the DIHANA corpus~\cite{Benedi2006} consists of 900 dialogs in Spanish between human speakers and a \ac{WoZ} telephonic train information system. The total number of annotated segments is 23,542, with 9,712 corresponding to user segments and 13,830 to system segments~\cite{Alcacer2005}. The set is partitioned into five folds to be used for cross-validation~\cite{Martinez-Hinarejos2008}. The dialog act annotations are hierarchically decomposed in three levels~\cite{Martinez-Hinarejos2002}. The first level represents the domain-independent intention of the segment, while the remaining are task-specific. In our experiments we focused on the first level, which has 11 different tags, out of which 5 are common to user and system segments.

\subsection{Classification Approach}

\begin{figure}[ht]
	\centering
	\includegraphics[width=\textwidth]{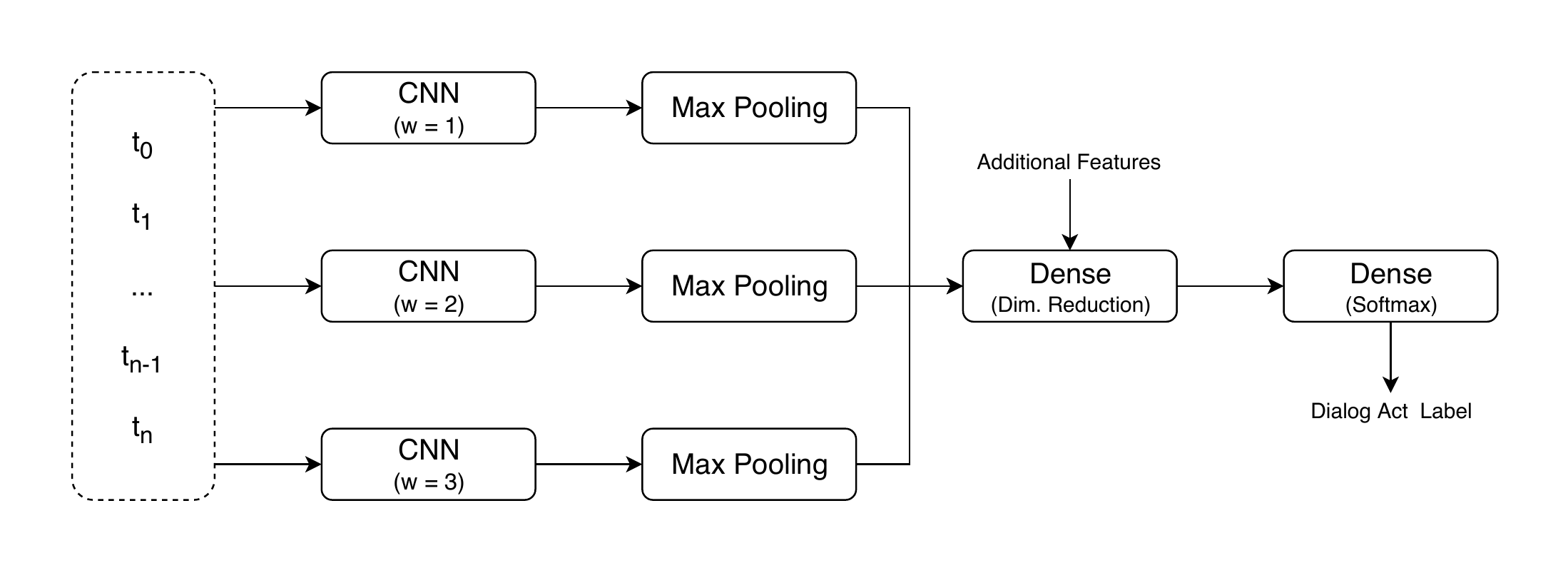}
	\caption{The generic architecture of the network used in our experiments. $t_i$ corresponds to the embedding representation of the $i$-th token. $w$ corresponds to the context window size of the \ac{CNN}. The number of parallel \acp{CNN} and the respective window sizes vary between experiments. Those shown in the figure correspond to the ones used by Liu et al.~\cite{Liu2017} in their experiments.}
	\label{fig:arch}
\end{figure}

As a classification approach, we adapted the state-of-the-art word-level approach by Liu et al.~\cite{Liu2017} to use characters instead of words as tokens. As shown in Figure~\ref{fig:arch}, the token embeddings are passed through a set of parallel temporal \acp{CNN} with different context window sizes followed by a max pooling operation. The results of those operations are then concatenated to form a representation of the segment. To achieve the state-of-the-art results, additional features concerning context information are appended to that representation before it is passed through a dimensionality reduction layer. Since our study focuses on the difference between using character- and word-level tokenization, we only included that information in a final experiment for comparison with the state-of-the-art. Finally, the reduced segment representation is passed through a dense layer with the softmax activation to obtain its classification.

\begin{figure}[ht]
	\centering
	\includegraphics[width=\textwidth]{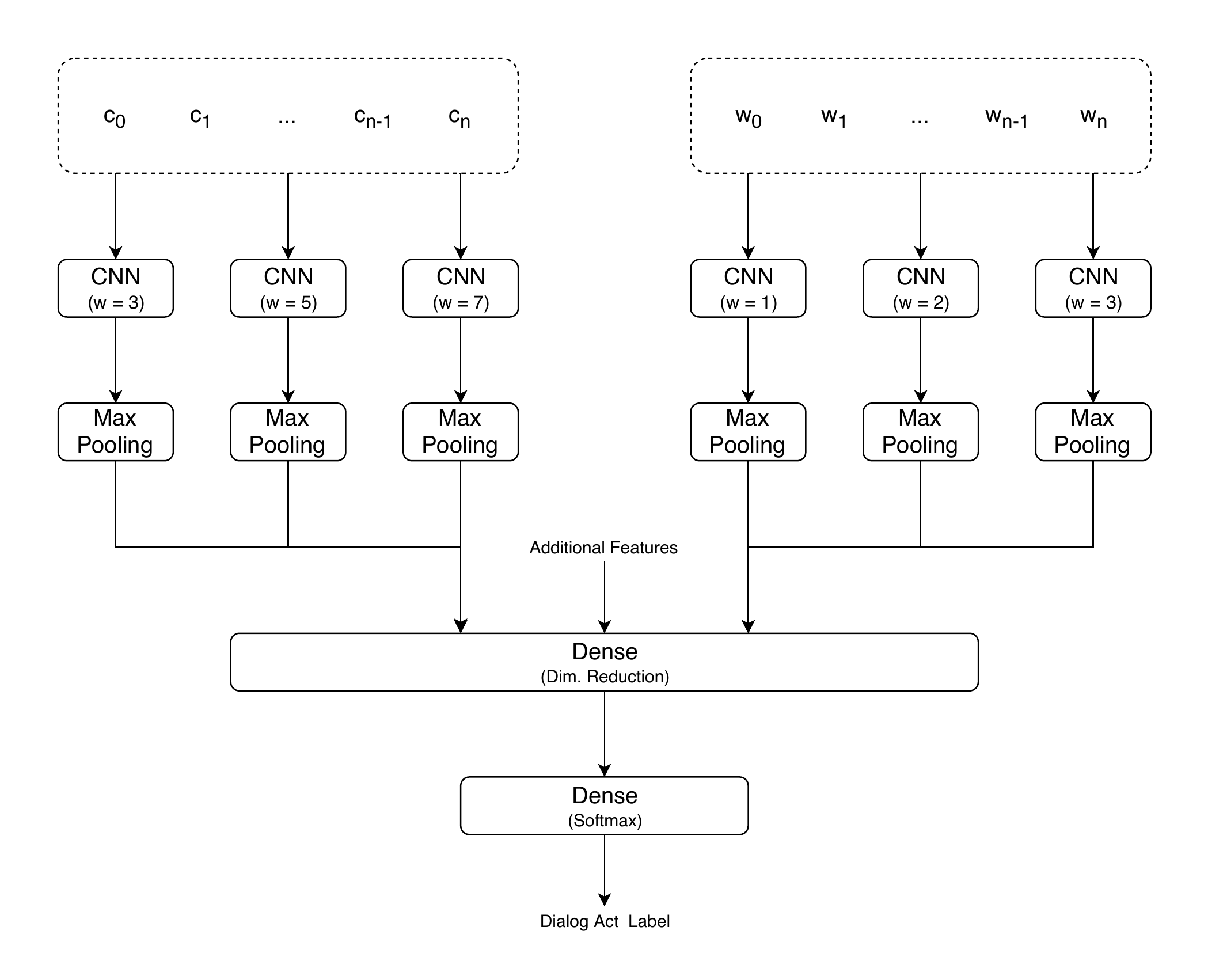}
	\caption{The architecture of the network that combines the character- and -word-level approaches. $c_i$ corresponds to the embedding representation of the $i$-th character while $w_i$ corresponds to the embedding representation of the $i$-th word. The context window sizes of the \acp{CNN} in the character-level branch refer to those that achieved best performance in our experiments. Those on the word-level branch correspond to the ones used by Liu et al.~\cite{Liu2017} in their experiments.}
	\label{fig:mergearch}
\end{figure}

In order to assess whether the character- and word-level approaches capture complementary information, we also performed experiments that combined both approaches. In that scenario, we used the architecture shown in Figure~\ref{fig:mergearch}. In this case, two segment representations are generated in parallel, one based on the characters in the segment and other on its words. Those representations are then concatenated to form the final representation of the segment. The following steps do not differ from the architecture with a single branch. That is, context information can be added to the segment representation before it is passed to the two dense layers. 

We used Keras~\cite{Chollet2015} with the TensorFlow~\cite{Abadi2015} backend to implement the networks. The training phase stopped after 10 epochs without improvement on the validation set. The results presented in the next section refer to the average ($\mu$) and standard deviation ($\sigma$) accuracy values over 10 runs. 

\subsection{Baselines}

In order to assess the performance of the character-level approach, in comparison to the word-level approach, we defined two baselines. One of them uses randomly initialized word embeddings that are adapted during the training phase, while the other uses fixed pre-trained embeddings. The latter were obtained by applying Word2Vec~\cite{Mikolov2013} on the English Wikipedia\footnote{\url{https://dumps.wikimedia.org/enwiki/}} and the Spanish Billion Word Corpus~\cite{Cardellino2016}. Additionally, we defined a third baseline that replicates the state-of-the-art approach by Liu et al.~\cite{Liu2017}. It consists of the baseline with pre-trained embeddings combined with context information from three preceding segments in the form of their gold standard annotations and speaker change information in the form of a flag. Similarly to Liu et al.~\cite{Liu2017}, we used three parallel \acp{CNN} with context window sizes one, two, and three in all the baselines.

%
%
%
%

\section{Results}
\label{sec:results}

Starting with the word-level baselines, in Table~\ref{tab:resbaseline} we can see that, in comparison to using randomly initialized word embeddings, using fixed pre-trained embeddings led to an average accuracy improvement of .64 and .88 percentage points on the validation (SWBD-V) and test (SWBD-T) sets of the Switchboard corpus, respectively. However, that was not the case on the DIHANA corpus, where the improvement was negligible. This can be explained by the difference in the nature of the dialogs between corpora. Since the Switchboard dialogs have a large variability in terms of style and domain, the performance on the validation and test sets is impaired when overfitting to the training data occurs. On the other hand, since most DIHANA dialogs are similar, the cross-validation performance is not impaired and may actually benefit from it. The improvement provided by context information is in line with that reported by Liu et al.~\cite{Liu2017}. Our results on the Switchboard corpus vary from those reported in their paper mainly because they did not use the standard validation and test partitions. 

\begin{table} [ht]
\caption{Accuracy results of the word-level baselines.}
\label{tab:resbaseline}
\begin{center}
    \begin{tabular}{l c c c c c c}
        \cline{2-7}
                              & \multicolumn{2}{c}{SWBD-V} & \multicolumn{2}{c}{SWBD-T} & \multicolumn{2}{c}{DIHANA} \tabularnewline
					          & $\mu$ & $\sigma$ & $\mu$ & $\sigma$ & $\mu$ & $\sigma$ \tabularnewline
        \hline
        Random                & .7617 &    .0019 & .7223 &    .0020 & .9196 &    .0013 \tabularnewline
        Pre-trained           & .7681 &    .0032 & .7311 &    .0026 & .9198 &    .0012 \tabularnewline
        Pre-trained + Context & .8129 &    .0030 & .7835 &    .0036 & .9826 &    .0004 \tabularnewline
    \end{tabular}
\end{center}
\end{table}

Regarding the character-level experiments, in Table~\ref{tab:reswindow} we can see that, as expected, considering each character individually is not the appropriate approach to capture intention. By considering pairs of characters, the performance improved by over 5 percentage points on both corpora. Widening the window up to five characters leads to a nearly 3 percentage point improvement on the Switchboard corpus, but less than 1 percentage point on the DIHANA corpus. However, it is important to note that while the results are above 90\% accuracy on the DIHANA corpus, they are below 80\% on the Switchboard corpus. Thus, improvements are expected to be less noticeable on the first. Using a window of seven characters still improves the results on the Switchboard corpus, but is not relevant on the DIHANA corpus. Considering wider windows is harmful on both corpora. However, note that, similarly to what Liu et al.~\cite{Liu2017} have shown at the word level, different context windows are able to capture complementary information. Thus, it is beneficial to combine multiple windows. In our experiments, the best results on both corpora were achieved using three context windows, which considered groups of three, five, and seven characters, respectively. The sizes of these windows are relevant, since the shortest window is able to capture most affixes in English and the small affixes in Spanish, the middle window is able to capture the larger Spanish affixes and most lemmas in both languages, and the widest window is able to capture larger words and inter-word information. Finally, it is relevant to note that the results on the DIHANA corpus are already above the word-level baselines.

\begin{table} [ht]
\caption{Accuracy results using different token context windows.}
\label{tab:reswindow}
\begin{center}
    \begin{tabular}{r c c c c c c}
        \cline{2-7}
                       & \multicolumn{2}{c}{SWBD-V} & \multicolumn{2}{c}{SWBD-T} & \multicolumn{2}{c}{DIHANA} \tabularnewline
        Window Size(s) & $\mu$ & $\sigma$ & $\mu$ & $\sigma$ & $\mu$ & $\sigma$ \tabularnewline
        \hline
                     1 & .6542 &    .0017 & .6081 &    .0023 & .8571 &    .0029 \tabularnewline
                     2 & .7221 &    .0047 & .6752 &    .0054 & .9154 &    .0014 \tabularnewline
                     3 & .7432 &    .0055 & .7000 &    .0035 & .9217 &    .0010 \tabularnewline
                     4 & .7456 &    .0019 & .7064 &    .0049 & .9222 &    .0014 \tabularnewline
                     5 & .7509 &    .0052 & .7091 &    .0038 & .9228 &    .0011 \tabularnewline
                     7 & .7535 &    .0023 & .7086 &    .0034 & .9224 &    .0013 \tabularnewline
                    10 & .7510 &    .0036 & .7097 &    .0035 & .9216 &    .0013 \tabularnewline
             (3, 5, 7) & .7608 &    .0033 & .7208 &    .0042 & .9244 &    .0012 \tabularnewline
    \end{tabular}
\end{center}
\end{table}

In Section~\ref{sec:character}, we hypothesized that capitalization is not relevant for dialog act recognition. In Table~\ref{tab:respre}, we can can see that the hypothesis holds for the Switchboard corpus, as the results obtained when using capitalized segments do not significantly differ from those obtained using uncapitalized segments. However, on the DIHANA corpus, using capitalized segments led to an average improvement of 1.81 percentage points. Since this was not expected, we looked for the source of the improvement. By inspecting the transcriptions, we noticed that, contrarily to user segments, the system segments do not contain mid-segment capitalization. Thus, proper nouns, such as city names which are common in the dialogs, are capitalized differently. Since only 5 of the 11 dialog acts are common to user and system segments, identifying its source reduces the set of possible dialog acts for the segment. Thus, the improvement observed when using capitalization information is justified by the cues it provides to identify whether it is a user or system segment. 

\begin{table} [ht]
\caption{Accuracy results using different segment preprocessing approaches.}
\label{tab:respre}
\begin{center}
    \begin{tabular}{l c c c c c c}
        \cline{2-7}
                                 & \multicolumn{2}{c}{SWBD-V} & \multicolumn{2}{c}{SWBD-T} & \multicolumn{2}{c}{DIHANA} \tabularnewline
					             & $\mu$ & $\sigma$ & $\mu$ & $\sigma$ & $\mu$ & $\sigma$ \tabularnewline
        \hline
        Capitalized              & .7604 &    .0028 & .7194 &    .0026 & .9425 &    .0015 \tabularnewline
        Punctuated               & .7685 &    .0021 & .7317 &    .0032 & .9371 &    .0007 \tabularnewline
        Capitalized + Punctuated & .7673 &    .0025 & .7314 &    .0040 & .9548 &    .0004 \tabularnewline
        Lemmatized               & .7521 &    .0027 & .7140 &    .0012 & .9239 &    .0006 \tabularnewline
    \end{tabular}
\end{center}
\end{table}

In Table~\ref{tab:respre}, we can also see that, as expected, punctuation provides relevant information for the task, improving the performance around 1 percentage point on both corpora. Using this information, the character-level approach surpasses the randomly initialized word-level baseline and is in line with the one using pre-trained word embeddings on the Switchboard corpus. Also expectedly, the decrease in performance observed when using lemmatized segments proves that affixes are relevant. However, that decrease is not drastic, which suggests that most information concerning intention can be transmitted using a simplified language that does not consider variations of the same lexeme and that those variations are only relevant for transmitting some specific intentions.

\begin{table} [ht]
\caption{Accuracy results using the combination of word and character-level representations.}
\label{tab:rescomb}
\begin{center}
    \begin{tabular}{l c c c c c c}
        \cline{2-7}
                              & \multicolumn{2}{c}{SWBD-V} & \multicolumn{2}{c}{SWBD-T} & \multicolumn{2}{c}{DIHANA} \tabularnewline
				              & $\mu$ & $\sigma$ & $\mu$ & $\sigma$ & $\mu$ & $\sigma$ \tabularnewline
        \hline
        Char + Word           & .7800 &    .0016 & .7401 &    .0035 & .9568 &    .0003 \tabularnewline
        Char + Word + Context & .8200 &    .0027 & .7901 &    .0016 & .9910 &    .0004 \tabularnewline
    \end{tabular}
\end{center}
\end{table}

Finally, Table~\ref{tab:rescomb} shows the results obtained by combining the word and character-level approaches. We can see that the performance increases on both corpora, which means that both approaches are able to capture complementary information. This confirms that information at the sub-word level is relevant for the task. When using context information, the combination of both approaches leads to results that surpass the state-of-the-art word-level approach on the Switchboard corpus by around .7 percentage points and to a nearly perfect score on the DIHANA corpus. Concerning the latter, it is not fair to compare our results with those of previous studies, since the only one that focused on Level 1 labels did not rely on textual information~\cite{Tamarit2008}.   

%
%
%
%

\section{Conclusions}
\label{sec:conclusions}

We have shown that there is important information for dialog act recognition at the sub-word level which cannot be captured by word-level approaches. We used character-level tokenization together with multiple context windows with different sizes in order to capture relevant morphological elements, such as affixes and lemmas, as well as long words and inter-word information. Furthermore, we have shown that, as expected, punctuation is important for the task since it is able to provide both direct and indirect cues regarding intention. On the other hand, capitalization is irrelevant under normal conditions. Finally, our experiments revealed that the character- and word-level approaches capture complementary information and, consequently, their combination leads to improved performance on the task. In this sense, by combining both approaches with context information we achieved state-of-the-art results on the Switchboard corpus and a nearly perfect score on the DIHANA corpus.

It is important to note that while one of the corpora used in our experiments features variable-domain human-human interactions in English, the other features fixed-domain interactions in Spanish between a \ac{WoZ} dialog system and its users. Thus, the importance of information at the sub-word level is not domain-dependent and it is not limited to a single language.

In terms of morphological typology, although English has a more analytic structure than Spanish, both are fusional languages. Thus, as future work it would be interesting to assess whether the conclusions of this study hold for analytic languages, such as Chinese, and agglutinative languages, such as Turkish.

%
%
%
%
%
\bibliographystyle{splncs04}
\bibliography{references}
\end{document}